# When Should an AI Act? A Human-Centered Model of Scene, Context, and Behavior for Agentic AI Design

Soyoung Jung[1], Daehoo Yoon[1], Sung Gyu Koh[2],
Young Hwan Kim[2], Yehan Ahn[2], and Sung Park[1]*

[1]Taejae Human-Centric AI Center (HCAC),
Taejae University, Seoul, Republic of Korea
[2]LG Electronics, Seoul, Republic of Korea
*Corresponding author: sjp@taejae.ac.kr

Agentic AI increasingly intervenes proactively by inferring users' situations from contextual data yet often fails for lack of principled judgment about when, why, and whether to act. We address this gap by proposing a conceptual model that reframes behavior as an interpretive outcome integrating *Scene* (observable situation), *Context* (user-constructed meaning), and *Human Behavior Factors* (determinants shaping behavioral likelihood). Grounded in multidisciplinary perspectives across the humanities, social sciences, HCI, and engineering, the model separates what is observable from what is meaningful to the user and explains how the same scene can yield different behavioral meanings and outcomes. To translate this lens into design action, we derive five agent design principles (behavioral alignment, contextual sensitivity, temporal appropriateness, motivational calibration, and agency preservation) that guide intervention depth, timing, intensity, and restraint. Together, the model and principles provide a foundation for designing agentic AI systems that act with contextual sensitivity and judgment in interactions.

## 1 INTRODUCTION

Agentic AI is characterized by its capacity to infer users' situations and needs and to plan and make decisions autonomously based on these inferences [1]. Moving beyond reactive systems, agentic AI aims to plan and act proactively without explicit user requests [2]. As such systems increasingly intervene on behalf of users, a critical design question emerges: not only how agents understand situations, but when, why, and whether they should act within them.

Despite substantial recent advances in agentic AI systems, a fundamental challenge remains: interpreting users' dynamically changing goals [3], [4], [5] while maintaining contextual coherence despite access to abundant knowledge and data [6]. This limitation suggests that agents often struggle not with sensing information, but with grasping the meaning underlying users' behaviors. In practice, agents face difficulties in integrating an understanding of the objectively observable situation (*Scene*), how that situation is subjectively experienced and interpreted by the user (*Context*), and how behavioral decisions are formed (*Human Behavior Factors*).

Existing psychology-based behavioral models have primarily explained human behavior by focusing on internal, micro-level mechanisms such as habits, decision-making processes, and cognitive biases. However, these models are limited in their ability to account for the concrete situations in which behavior unfolds and the dynamically changing contexts surrounding it. This study proposes an integrative conceptual model that considers behavior not only in terms of internal individual factors, but also in relation to the situation in which behavior unfolds (*Scene*), the contextual meaning assigned to that situation (*Context*), and the factors through which behavior becomes likely or unlikely (*Human Behavior Factors*). Drawing on perspectives from the humanities and social sciences as well as engineering approaches, we present the model in two directions: (1) identifying the components of *Scene, Context, and Human*

*Behavior Factors* required to interpret user behavior, and (2) explaining how behavior emerges through processes of intention and choice.

## 2 RECONCEPTUALIZING HUMAN BEHAVIOR FOR AGENTIC AI

This study synthesizes multidisciplinary perspectives to reconceptualize human behavior for agentic AI design, with a focus on how situational meaning is constructed and interpreted. Human behavior should not be understood as a single, indivisible unit, but rather as a process structured across multiple levels. This hierarchical organization of behavior parallels the brain's information-processing mechanisms [7], which range from reactive reflexive processing to higher-order meaning-based representations [7], [8], [9], [10], [11].

These levels can be illustrated through a single everyday example: eating. At the low level, behavior consists of simple motor actions, such as finger movements or chewing. At the mid level, behavior involves goal-directed actions, such as picking up food or holding a spoon. At the high level, these actions are integrated into a meaningful activity, where eating is understood not merely as a sequence of movements, but as an activity shaped by context and meaning.

For agentic AI, the primary level of interest is this high-level form of behavior. Although terminology varies across studies, we refer to this level as *Activity*. However, much of the existing research treats this high level merely as activity [8], without sufficiently accounting for the role of context and subjective meaning. In practice, an *Activity* varies depending on with whom it is performed, where it takes place, and what objects are involved, as well as on the subjective meaning it holds for the individual. Accordingly, *Activity* is defined as the perceptible unit of behavior and serves as an anchor for analyzing and predicting behavior in situated contexts. Based on this framing, we discuss three key characteristics that influence the manifestation of behavior: (1) *Scene*, (2) *Context*, and (3) *Human Behavior Factors*.

### 2.1 Key Characteristics of Human Behavior

*2.1.1 Scene*

We reviewed two representative perspectives: scene understanding in vision and robotics, which perceives and analyzes scenes from a third-person viewpoint, and scenario-based design in HCI, which aims to understand and design situations as a whole. In scene understanding, a scene is defined as a snapshot captured at a specific moment in time. Within this snapshot, elements such as the background, objects, agents, activities, and the relationships among them are jointly represented [12], [13]. Although scenario-based design does not explicitly use the term "scene," it conceptualizes comparable units as episodes of human action in which contextual elements are integrated for design purposes [14], [15], [16]. *Scene*s are recognized or designed with a primary focus on action and are approached through objectively identifiable elements, such as the surrounding environment, the actors involved, and the sequence of actions being performed. Accordingly, a scene can be characterized as a configuration of objectively identifiable elements that include observable actions.

*2.1.2 Context*

Perspectives from neuroscience, which explain context in terms of how users receive and interpret information, and from HCI, which approaches context from an observer's standpoint to capture and design users' contextual information, were reviewed. In neuroscience, context is defined as the overall situation surrounding an event, and prior work explains the factors that shape how humans perceive and process information within that situation [17]. In HCI, context is commonly defined as any information that can be used to characterize the situation of an entity [18]. Although these two fields differ in perspective -- focusing respectively on the first-person user and the third-person observer -- they share a common emphasis on meaning construction as a basis for understanding and interpreting situations. For example, *Context* determines whether a given space is experienced as one's own home or a friend's home, or whether a particular moment is interpreted as time for leisure or time for sleep. In this way, context provides subjective meaning to a situation. A key characteristic of *Context* is that it consists of elements through which situations are subjectively interpreted and imbued with meaning.

*2.1.3 Human Behavior Factors*

Discussions on what determines behavior have extensively examined internal factors, external factors, and interactional factors. Among these, three representative psychology- and HCI-based theories are examined with a focus on behavioral explanation and prediction: the Theory of Planned Behavior (TPB) [19], the COM-B model [20], and a behavior model for persuasive design [21]. Although these theories differ in how they conceptualize and explain behavior, this study extracts and organizes the conceptual factors that are considered essential for predicting behavior. Specifically, these factors include *Attitudes* toward the behavior, *Opportunities* and *Capabilities* that enable or constrain action, *Motivations* that drive behavior, and *Triggers* that initiate action.

## 2.2 Relationships of Scene, Context, Human Behavior Factors

Human behavior can be understood in an integrated manner through three core concepts: *Scene, Context, and Human Behavior Factors*. A *Scene* consists of objectively identifiable elements; however, it does not inherently capture the meaning those elements hold for an individual. *Context*, therefore, plays the role of assigning meaning to the objective elements of a *Scene*. For example, the same living-room scene can mean "my home" or "someone else's," leading to different interpretations and actions.

The process through which actual behavior emerges also involves factors that determine behavior. However, because these factors belong to the domain of individual cognition and perception, they are difficult to observe directly or to identify in a clear and explicit manner. Rather than serving as observable variables for directly predicting behavior, these *Human Behavior factors* function as conceptual criteria for judging which behaviors are more likely to occur within a given *Scene* and *Context*. *Human Behavior Factors* thus shape the direction of behavioral likelihood in a given situation. For instance, among these factors, *Capability* involves a cognitive process through which individuals assess whether an action is physically or cognitively feasible, based on the meaning constructed from the *Scene* and *Context*.

Taken together, behavior can be understood as an outcome formed through the combination of *Scene*, *Context*, and *Human Behavior Factors*. Behavior is defined as the result of integrating subjectively constructed meaning with an objectively observable situation. Rather than being directly observable, it represents a meta-level construct that is formed through interpretation and prediction. Behavior is defined as human activity that is imbued with meaning through *Context* within a given situation (*Scene*).

## 3 A CONCEPTUAL MODEL OF HUMAN BEHAVIORAL PROCESSES

Based on the key characteristics that influence human behavior, we propose a conceptual model for a deeper understanding of human behavior. The model starts from an observable *Scene*, moves through *Context*, which encapsulates subjective meaning, and *Human behavior factors*, which shape behavioral possibilities, and ultimately explains the process through which behavior emerges (see Figure 1).

The components of a *Scene* include *Actor*(s); movable *Objects*; the *Background*, which consists of fixed surfaces and structures such as the ground, floor, and walls; and *Activity*, which refers to a sequence of actions performed by an actor. *Activity* serves as the anchor of behavior. Based on these observed elements of a *Scene*, *Context* assigns meaning to each component. Because the elements of a *Scene* are objectively identifiable, they are positioned externally in the model.

*Context* and *Human Behavior Factors*, as illustrated by the boxed components, operate through internal cognitive and perceptual processes that lead to behavior generation (Internal Behavior Generation Loop).

The components of *Context* consist of several dimensions: *Spatial context*, related to layout, location, illumination, and noise; *Temporal context*, related to time of day, frequency, and duration; *Interoceptive context*, related to physiological signals such as hormones, heart rate, and respiration; *Individual context*, related to personal characteristics such as user profiles, interests, and preferences; and *Social/Cultural context*, related to social relationships, culture, and norms.

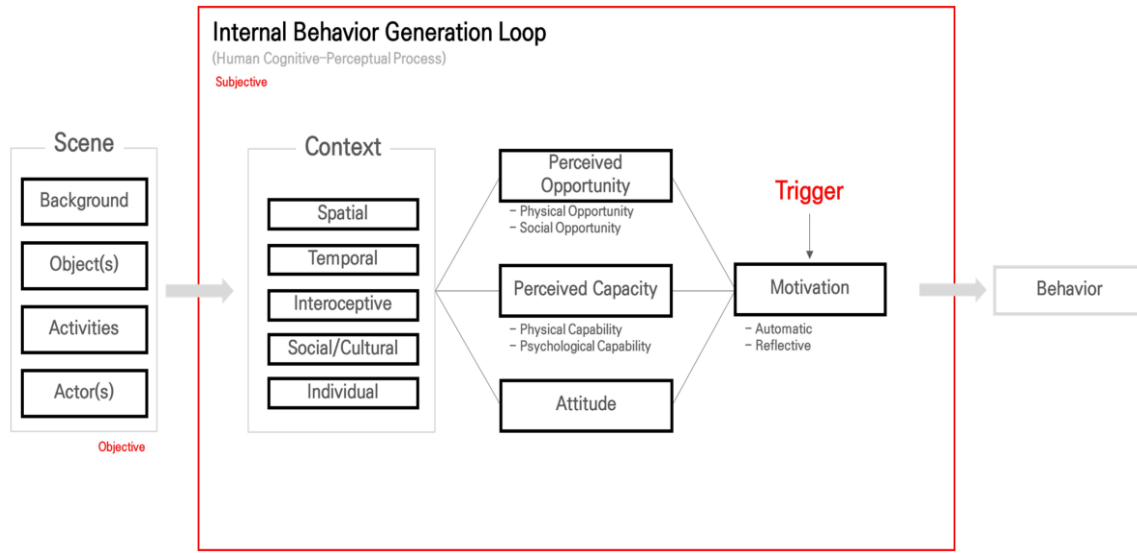

Figure 1: A Conceptual Model of Human Behavioral Processes

*Human Behavior Factors* consist of the following components. *Attitude* refers to an individual's evaluation of a behavior. *Perceived opportunity* denotes the resources in the external environment as perceived by the individual and is divided into physical (e.g., environment, time, finances, objects) and social (e.g., people, organizations, culture) opportunities. *Perceived capability* refers to the individual's perceived ability to perform a behavior and is divided into physical (e.g., body condition, stamina, strength) and psychological (e.g., comprehension, reasoning, affective states) capabilities. *Motivation* represents the internal drive that initiates behavior and is divided into automatic motivation (e.g., habits, instincts) and reflective motivation (e.g., evaluation and planning of behavior). Finally, a *Trigger* refers to a signal that activates the transition from motivation to actual behavior. *Attitude, Opportunity*, and *Capability* influence *Motivation*, and *Motivation* forms the foundation of behavior formation. However, *Motivation* alone does not always result in behavior; when a *Trigger* is present, motivation can be activated and translated into actual behavior.

Figure 2 illustrates the proposed conceptual model through an example drawn from an everyday situation. In this example, the user's situation of boarding a bus after work is presented as the *Scene*, while multiple contextual factors demonstrate how this *Scene* is interpreted. Through this interpretive process, the user's act of listening to music on a smartphone can be understood not merely as an activity, but as an automatic behavior chosen to seek brief relief in a post-work situation characterized by elevated stress. In other words, within the proposed model, behavior refers not only to what choice the user makes in a given situation, but also to the contextual reasons through which that choice is formed.

From the perspective of agentic AI design, the proposed model provides a structured basis for reasoning about intervention timing and appropriateness. By distinguishing *Scene, Context, and Human Behavior Factors*, an agent can assess not only what is happening, but whether the situation is interpretable, actionable, or should remain unaltered. This enables agents to differentiate between moments that call for proactive support, moments that require restraint, and moments where intervention may be inappropriate.

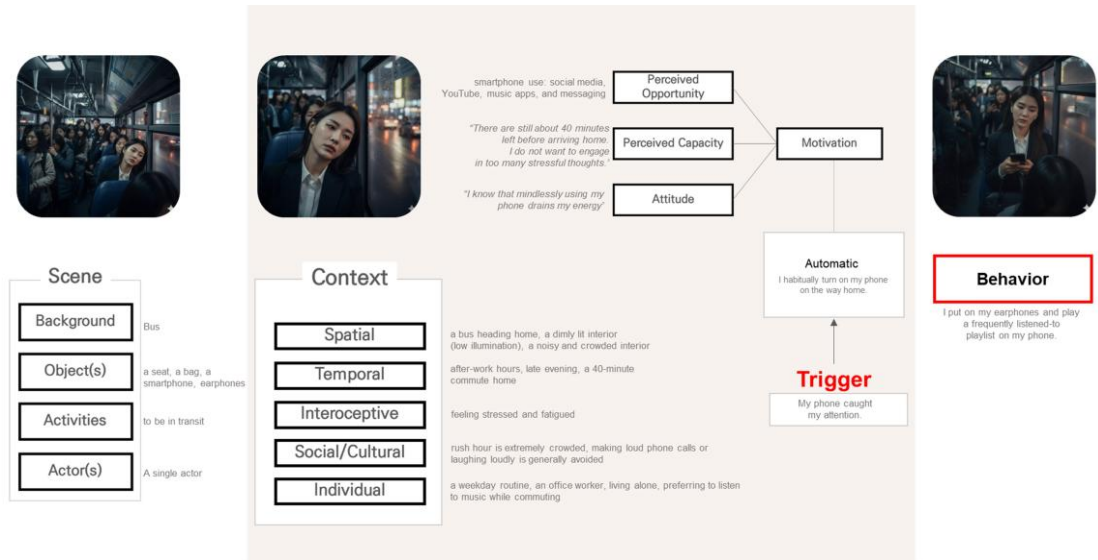

Figure 2: An Illustrative Example of Behavior Generation from Scene and Context

## 4 FROM MODEL OF DESIGN PRINCIPLES FOR AGENTIC AI

The conceptual model proposed in this study is not intended as an explanatory theory of behavior, but as a design lens for reasoning about when and how an agent should intervene in users' situations. To bridge the gap between conceptual understanding and design decision-making, we derive a set of agent design principles grounded in the relationships among *Scene, Context, and Human Behavior Factors*. These principles translate the model into actionable guidance for designing agentic AI interventions with contextual sensitivity, appropriate timing, and respect for user agency.

### 4.1 Behavioral Alignment Principle

Agent interventions should be aligned with what the user is doing. Human behavior operates at different levels, ranging from immediate actions to higher-level goals and meanings. Intervening at a level that does not match the user's activity can interrupt rather than support behavior. For example, suggesting a long-term goal change while a user is executing a simple task may feel disruptive. Based on the model's view of *Activity* as the core unit of behavior, this principle encourages agents to adjust the scope and abstraction of their interventions to fit the user's activity.

### 4.2 Contextual Sensitivity Principle

Whether an agent should intervene depends not only on what is happening, but on how the situation is experienced by the user. The same observable scene can carry different meanings depending on contextual factors such as time, place, personal state, and social norms. Interventions that rely only on objective cues risk feeling intrusive or irrelevant when they overlook this contextual meaning. This principle emphasizes that agents should adapt their intervention strategies based on spatial, temporal, individual, and social-cultural context, so that actions align with how the user interprets the situation.

### 4.3 Temporal Appropriateness Principle

Agent interventions should be guided not only by what to do, but by when to act. Even well-aligned and contextually appropriate interventions can feel disruptive if they occur at the wrong moment. Because human behavior unfolds over time through interpretation rather than as an immediate reaction to observable cues, agents must carefully judge timing. This principle emphasizes that agents should distinguish moments that invite proactive support from moments that call for waiting or restraint, allowing user behavior to proceed without unnecessary interruption.

### 4.4 Motivational Calibration Principle

Agent interventions should be adjusted in strength and form based on factors that influence whether a behavior is likely to occur. These factors include the user's attitudes, perceived opportunities, capabilities, motivation, and triggers within a given *Scene* and *Context*. Instead of treating intervention as a simple on-or-off decision, agents should vary how strongly and in what way they intervene. For example, agents may provide information when users lack capability, or modify the environment when opportunities are limited. This principle encourages agents to tailor their interventions to the user's behavioral readiness.

### 4.5 Agency Preservation Principle

Agentic AI systems should preserve users' sense of agency by maintaining user control over intervention outcomes. As agents become increasingly proactive, there is a risk that automated actions may override users' intentions or reduce their perceived autonomy. Grounded in the model's human-centered perspective, this principle emphasizes that interventions should remain assistive rather than substitutive. Agents should enable users to defer, reject, or modify suggested actions, ensuring that proactive support reinforces rather than diminishes the user's role as the primary decision-maker.

## 5 DISCUSSION

This study proposes a conceptual model for understanding human behavior in the design of agentic AI systems. Drawing on multidisciplinary perspectives from psychology, cognitive science, HCI, and robotics, the model aims to support agent behavior that is grounded not only in observable situations but also in how those situations are interpreted by users. The model makes two primary contributions. First, it explicitly distinguishes between *Scene, Context*, and *Human Behavior Factors* as complementary components for interpreting behavior. While recent agentic systems have advanced in sensing and inference, they often struggle to connect what is happening with what it means to the user. By separating objectively identifiable elements of a situation (*Scene*) from the elements that assign subjective meaning (*Context*), and by incorporating factors that shape behavioral likelihood (*Human Behavior Factors*), the model provides a structured account of why the same situation can lead to different behaviors across users.

Second, the model conceptualizes behavior as an outcome of interpretation, intention, and choice rather than as a direct response to observable cues. Through this perspective, agentic AI can move beyond surface-level action recognition toward reasoning about users' contextual states when deciding whether, how, and when to intervene. Prior work has highlighted the importance of context in behavior understanding and proactive systems [22], [23], [24]. Building on this trajectory, the present model emphasizes the need for agent design approaches that anticipate users' own interpretations of their behavior, rather than relying solely on external observation or behavioral logs.

This work is primarily conceptual and has not yet been empirically validated [25]. Future research should examine how specific components of Scene and Context contribute to behavior prediction and decision-making, and how the model can be operationalized using sensor data and interaction traces. Overall, this study advances agentic AI design toward a more human-centered perspective. Rather than framing agents as systems that simply recognize actions or automate tasks, the proposed model positions them as systems that reason about human behavior in light of situational meaning. By foregrounding context and interpretation as central design considerations, the model offers a foundation for developing agentic AI systems that act with greater sensitivity, restraint, and alignment with human judgment.

**ACKNOWLEDGEMENT**

This research was sponsored and supported by the collaboration project between LG Electronics, Inc. and Taejae University.